\documentclass[wcp]{jmlr}

\usepackage{tikz}
\usetikzlibrary{positioning}
\usetikzlibrary{shapes.multipart}
\usetikzlibrary{shapes.geometric}

\usepackage{amsmath,amssymb}
\usepackage{graphicx,url}
\usepackage{svg}
\usepackage{soul}

\jmlrvolume{266}
\jmlryear{2025}
\jmlrworkshop{Conformal and Probabilistic Prediction with Applications}
\jmlrproceedings{PMLR}{Proceedings of Machine Learning Research}

\newcommand{\Tk}{{\mathcal{T}^{(k)}}}
\newcommand{\D}{{\mathcal{D}}}

\newcommand{\gr}[1]{{\color{black} #1}}

\newcommand{\grCom}[1]{{ \color{red} GESINE COMMENT: #1}}

\newcommand{\andeCom}[1]{{ \color{red}  ANDREW COMMENT: #1}}

\firstpageno{1}

\begin{document}
\title{Individualised Counterfactual Examples Using Conformal Prediction Intervals}

\author{\Name{James M. Adams} \Email{jadams@turing.ac.uk}\\
      \addr{The Alan Turing Institute,
       London, NW1 2DB, United Kingdom.
       }\vspace{0.5ex}
       \AND
       \Name{Gesine Reinert}
\Email{reinert@stats.ox.ac.uk}\\
       \addr{
       Department of Statistics,
       University of Oxford,
       OX1 3LB, United Kingdom.} \vspace{0.5ex}
         \AND 
       \Name{Lukasz Szpruch} \Email{l.szpruch@ed.ac.uk}\\
       \addr{University Of Edinburgh, EH9 3FD, United Kingdom.
       }\vspace{0.5ex}
        \AND
       \Name{Carsten Maple} \Email{cm@warwick.ac.uk}\\
       \addr{University Of Warwick, 
        Coventry, CV4 7AL, United Kingdom.
       }\vspace{0.5ex}   \AND
       \Name{Andrew Elliott}
\Email{andrew.elliott@glasgow.ac.uk}\\
       \addr{School of Mathematics and Statistics,
       University of Glasgow,
       G12 8QQ, United Kingdom.
       }
       }
       
\editor{Khuong An Nguyen, Zhiyuan Luo, Tuwe L\"ofstr\"om, Lars Carlsson and Henrik Bostr\"om}

\maketitle
\begin{abstract}%
Counterfactual explanations for black-box models aim to pr  ovide insight into 
an algorithmic decision to its recipient. For a binary classification problem an individual 
counterfactual details which features might be changed for the model to infer the opposite class. 
High-dimensional feature spaces that are typical of machine learning classification models 
admit many possible counterfactual examples to a decision, and so it is important to identify additional criteria to select the most useful counterfactuals.

In this paper, we explore the idea that the counterfactuals should be maximally 
informative when
considering the knowledge of a  specific individual %
about the underlying classifier. 
To quantify this \gr{information gain} we explicitly model the knowledge of the individual, and assess the uncertainty \gr{of predictions which the individual makes} by  the \emph{width of a conformal prediction interval}. %
Regions of feature space where the prediction
interval is wide correspond to areas where the confidence in decision making is low, and an additional 
counterfactual example might be more informative to an individual.

{To explore and 
evaluate} 
our individualised conformal prediction interval counterfactuals (CPICFs),  %
{first we present a synthetic \gr{data set on a hypercube} %
which allows us to fully visualise the decision boundary, conformal intervals via three different methods, and resultant CPICFs. Second, in this synthetic \gr{data set} we explore the impact of a single CPICF on the knowledge 
of an individual locally around the original query. Finally, in both our synthetic \gr{data set} and a complex real world dataset with 
a combination of continuous and discrete variables, we measure the utility of these counterfactuals via data augmentation, testing the performance on a held out set. 
}
\end{abstract}

\vspace{2ex} %


\section{Introduction}

Machine learning algorithms are essential in the analysis of large volumes of tabular data such as 
loan applications, medical data or credit card transactions \citep{wachter_counterfactual_2017}. 
Consequential decisions in these areas are often made by %
high-performing black-box algorithms trained on large quantities of data. However an explanation of the output is legally required to satisfy General Data Protection Regulation (GDPR) requirements of an explanation of an algorithmic decision \citep{goodman_european_2017}, 
so that both users and recipients of the decisions trust the model. An explanation can also be used to demonstrate both fairness and transparency to customers. 

One possible approach is the use of so-called ``counterfactual explanations'' which give changes to an input instance that change the outcome of the %
black-box decision making system. For example, in the case of fraud detection they provide an alternative instance with suggested changes that might make a fraudulent transaction be labelled as non-fraudulent. There is a 
{substantial} 
literature on algorithms for the generation of counterfactual instances - see for example \cite{guidotti_counterfactual_2024} for a review. Importantly, for a particular 
input there are many possible counterfactuals, and so there are various criteria that can be used to choose between them (validity, actionability, sparsity, plausibility, and causality, see \cite{russell_learning_nodate}) as well as metrics that can measure these desirable features, see  \cite{verma_counterfactual_2024}. 

The aim of this paper is to use counterfactuals to improve %
{an individual's knowledge of a black box classifier}. Intuitively, suppose that {an individual has access to a set of inputs and outputs (including probabilities) of 
a black box binary classifier}. 
\gr{For a specific data point, the individual asks for a counterfactual from the organisation that holds the classifier. How should the organisation choose a counterfactual to convey}
the largest amount of information about the classifier \gr{based on the data which the individual  has access to}?

Adding counterfactuals is a particular instance of data augmentation. Data augmentation has been shown to be able to improve classification, see for example \cite{moreno2020improving} or  {\cite{wang2024comprehensive} for a survey.} 
Often these methods learn the overall data distribution and aim to generate new data that is consistent with the data but also adds diversity. %
Counterfactual data augmentation has been carried out for example in \cite{pitis2022mocoda}, which is a model-based approach. Often, as in \cite{temraz2022solving}, counterfactuals are generated to address a data set imbalance.

%

In \cite{altmeyer_faithful_2023}, \cite{lei_conformal_2021}, and \cite{chen_conformal_2024}, counterfactuals are generated taking conformal prediction sets into account, mostly for medical applications, with the assumption of i.i.d.\,observations. %
 Conformal prediction, going back to \cite{gammerman1998learning}, %
has become a key method for obtaining theoretical guarantees for prediction from black-box models. 
Conformal prediction gives a prediction interval for $Y_{n+1}$ associated with a new data point $X_{n+1}$, with guarantees obtained when $(X_{n+1},Y_{n+1}), (X_i, Y_i), i=1, \ldots, n$ are exchangeable.

We argue that, if one had a choice of where to put a  data point to obtain the largest amount of information about the classifier,  %
a point should be chosen that has a wide conformal prediction interval associated with it {in a model derived from the information that the individual holds}, and that has the opposite classification as a nearby data point that is already observed. Knowledge of this information should most reduce the uncertainty of the prediction. We therefore present a method of generating counterfactuals for a binary classification problem by using the inverse width of conformal prediction intervals %
as a measure of the informativeness of a counterfactual. This method is evaluated for applications in data augmentation of tabular data,  
%
%
{%
and assessed by examining the enhancement to the} 
individual's  understanding. 

This paper is structured as follows. Sec.~\ref{sec:lit} briefly reviews relevant literature from the fields of counterfactuals and conformal prediction. Sec.~\ref{sec:method} details our proposed counterfactual generation method. In Sec.~\ref{sec:exp} the method is illustrated on a specifically designed synthetic data set, and on a synthetic benchmark data set from \cite{padhi_tabular_2021}. %
Conclusions are drawn in Sec.~\ref{sec:concl}.
The code is available at \href{https://github.com/alan-turing-institute/CPICF}{https://github.com/alan-turing-institute/CPICF}.%
\section{Related literature on counterfactual generation} \label{sec:lit}

Early work on generating a counterfactual example, $X'$, to an original sample point $X_i$ used a minimisation algorithm to 
change prediction class, and a distance function $d(X', X_i)$ to induce sparse {changes} and proximity to the original 
example \citep{wachter_counterfactual_2017}. More recent work has developed algorithms that incorporate other desirable 
features such as the creation of a diverse collection of counterfactuals. 
This offers the user options of how the output might be changed, and increases the chances that one of the options is actionable. DiCE (Diverse 
Counterfactuals) \citep{mothilal_explaining_2020} incorporates proximity, diversity and user constraints (e.g. to ensure that a counterfactual is actionable) into its loss function. The software package has both a genetic loss minimisation, and a random search for counterfactual generation\footnote{\href{https://github.com/interpretml/DiCE}{https://github.com/interpretml/DiCE}}. DiCE has been examined for counterfactual generation for ATM fraud detection models in the literature \citep{vivek_explainable_2024}. ALIBI\footnote{\href{https://github.com/SeldonIO/alibi}{https://github.com/SeldonIO/alibi}} 
is a technique based on a training a surrogate reinforcement learning algorithm to produce the required counterfactual examples on the black box model of interest \citep{samoilescu_model-agnostic_2021}. Other algorithms, such as CLUE %
described in \cite{antoran_getting_2021}, generate counterfactuals that aim to minimise epistemic uncertainty (pertaining to the model), and so provide examples of model features that might be changed to reduce uncertainty.

\cite{nguyen_efficient_2022} have demonstrated how counterfactuals might be used to augment tabular data (Classification with Counterfactual Reasoning and Active Learning -- CCRAL). 
Suppose  a binary classifier  has been trained on a limited set of data produces prediction scores for the class label. 
For some samples the class prediction probabilities are close to $0.5$, indicating some uncertainty in their labelling. Counterfactuals are generated by changing the treatment feature of interest, and then labelling the new counterfactual 
point using its nearest real sample. The authors demonstrate improved performance using this data augmentation protocol.

In \cite{la2025novel}, a data augmentation strategy for credit scoring modelling  is proposed by adding synthetic instances obtained along the decision boundary of the model. They employ the framework CASTLE from \cite{la2021castle}, inputting the set of misclassified samples to CASTLE to obtain a vector which indicates how the input data should be perturbed in order to increase its probability of being correctly classified. %

An alternative approach to generating counterfactuals is based on
conformal prediction, as %
in \cite{altmeyer_faithful_2023}. Conformal prediction makes use of
{an additional}
calibration set %
to produce a prediction with a predetermined error (or miscoverage) %
$\alpha$. %
Larger prediction sets correspond to points with a higher prediction uncertainty. In their {\emph{Energy-Constrained Conformal Counterfactuals} method,} \cite{altmeyer_faithful_2023} include the prediction set size as a penalty in their objective function, which encourages the generation of counterfactuals with lower uncertainty. %
They seek low predictive uncertainty assuming a global model and do not model \gr{an individual's} knowledge.

\gr{We end this survey with a caveat}: as for example \cite{veitch2021counterfactual} observe \gr{by comparing their method to other methods}, \gr{of course} adding a counterfactual may not necessarily improve prediction \gr{if it is not well chosen}.

\section{Counterfactual generation method} \label{sec:method}

The general scenario in this paper is that 
data $\mathcal{D}\gr{= \{(X_i,Y_i),  i=-t,\ldots n\}}$ are available to an entity (a bank, say),
split into training data $\gr{\mathcal{T}= \{(X_i,Y_i),  i=-t,\ldots 0\}}$, on which a model is trained which predicts an outcome $Y$ for an observed $X$ \gr{from an arbitrary space $\mathcal{X}$, such as a space of financial transactions,} and calibration data $\gr{\{(X_i,Y_i),  i=1,\ldots n\}}$ {used as part of conformal prediction framework}.
Here we focus on  \gr{binary} outcomes $Y_i$.
For binary classification, %
\gr{a} (usually) black-box classification model $h_\theta = h_{\theta (\mathcal{D})}: \mathcal{X} \rightarrow \{0,1\} $ with parameters $\theta$, %
has been trained 
\gr{on $\mathcal{T}$}
to obtain classifications
$\hat{Y}_i =h_\theta (X_i) \in \{0,1\}$, {for $i=1, \ldots n$}. %
However, often underlying the classification prediction is a score, or an estimated probability $\gr{\mathbb{P}_\theta( Y_i = 1{| X_i}) =} p_\theta(X_i) \in [0,1]$, %
{%
on
\gr{if} an given $X_i$}, 
where 
$p_\theta = p_{\theta ({\mathcal{T}})}: \mathcal{X} \rightarrow [0,1] $,  
 with parameter $\theta$,  has been trained \gr{on $\mathcal{T}$.}
 \gr{From $p_\theta (X_i) $ finally a binary classification is derived, usually by thresholding}. 
\gr{Our} aim is to generate a counterfactual, denoted $X'\gr{=X'(X)}$, for \gr{an} observation $X$; thus, $X'$ has the opposite classification to $X$. 

\subsection{Modelling an individual's knowledge}
\label{sec:modelKnowledge}

\gr{A key novelty of this paper, and a}n important part of this novel counterfactual framework,  is to explicitly model
the knowledge of each individual and tailor the counterfactual examples to each
individual. For example, in a loan context, %
suppose that an individual has successfully applied for
multiple 
loans of  $ \pounds 1000$ and was refused a loan of $\pounds 2000$. \gr{The individual would like to understand why the second application was refused, rather than why the first application was successful.} In
this situation, instead of creating a counterfactual of $ \pounds 1000$,  %
it would be more informative for the 
individual 
to \gr{obtain a counterfactual closer to $ \pounds 2000$, with a change of some features such as loan purpose or credit rating.} 

To model the viewpoint of an individual, we suppose that an individual $k$  has access to a
subset $\mathcal{T}^{(k)} $ of the full {training} 
dataset \gr{$\mathcal{T}$} {and the probabilities or scores of these points in the full black box classifier - the latter of which is a somewhat strong assumption which could be relaxed in future work}. %
In the loan example, %
$\Tk$ could be %
the set of
loan applications that the individual %
submitted, {with corresponding application scores}. %
For simplicity %
we %
assume that the individual uses
the same model class as the original classifier to model their knowledge (we
leave the option of different classifier choices to future work).  

\gr{
The counterfactual example is generated by the entity based on the scores of the points in the calibration data set.
Due to the sensitivity of the function $p_\theta$ (at points not seen in the data set),  to generate a counterfactual for  individual $k$ that only has access to a subset $\Tk$ of the training data $\mathcal{T}$, the counterfactual example is created not based on $\mathcal{T}$ but only based on $\Tk.$} 
We  model the individual's
knowledge of the classifier in the same way as we would model the original
classifier $h_\theta$ \gr{trained on $\mathcal{T}$, but now trained on $\Tk$,} namely as $%
{h}_{\theta{(\Tk)}}$,  which for
notational convenience we refer to as $%
{h}_{\theta_{k}}$, \gr{obtained via thresholding from $p_{\theta_k}$, classifying outcomes as $1$ when $p_{\theta_k}$ exceeds a given threshold}. 
\gr{The entity then 
chooses a counterfactual example which takes the uncertainty in the prediction model of individual $k$ into account, by employing conformal prediction intervals.}

Alternatives, such  as using the individual's own classifier probabilities for the target of a regression model,  
{will be considered in future work.}

\subsection{Conformal prediction intervals}
\label{sec:conformal}

Conformal classifiers produce a prediction set, whose size enables it to meet the 
required prediction error 
$\alpha$. Here is a brief description of split conformal prediction, going back to \cite{gammerman1998learning}.
Suppose that a model is trained on data $(X_i, Y_i), i=-t, \ldots, 0$. %
Given a calibration set of data points $Z_i = (X_i, Y_i)$, $i = 1, \dots, n$, with $X_i \in  \mathcal{X}, Y_i \in  \mathcal{Y}$,  a (possibly black box) model $\hat{f}: \mathcal{X} \rightarrow \mathcal{Y}$, %
which is fitted using a training data set, produces predictions for $Y \in  \mathcal{Y}$ given $X\in  \mathcal{X}$. A (non-conformity) score function $S: \mathcal{X} \times \mathcal{Y} \rightarrow \mathbb{R}$ judges these predictions, the smaller the score, the better (in a regression context one could think of residuals as scores; for more details see for example \cite{papadopoulos_normalized_2008} and \cite{lei_distribution-free_2018}). What is observed are the scores $S(X_i, Y_i)$. These scores are then used to obtain a prediction set for the  next test point. The idea is that when we see $X$ we predict a $Y$ for which the score $S(X,Y)$ is ``typical''. 
Mathematically, 
we compute the score for each calibration data-point $S_i = S(X_i, Y_i)$, and take the order statistics $S_{1:n} \le S_{2:n} \le \cdots \le S_{n:n}$. For a  $1 - \alpha$-prediction set for $Y$ based on an observed $X$ we  find 
$$\hat{q} = S_{i:n} \text{ where } i = i_n = \lceil (1-\alpha)(n+1) \rceil,$$
and we use as the prediction set
\begin{align} \label{predset}
    \widehat{C}_{\alpha, n}\left(X\right)=\left\{y \in \mathcal{Y}: S\left(X, y\right) \leq \hat{q}\right\},
\end{align}
of width, or size, $C_{\alpha, n}(X) = |  \widehat{C}_{\alpha, n}|.$ 
For exchangeable data  we have the coverage guarantee 
\begin{align} \label{eq:covguarantee}
      1-\alpha \leq \mathbb{P}\left(Y \in \widehat{C}_{\alpha,n } (X)\right)  = { \lceil (1-\alpha)(n+1) \rceil}({n+1})^{-1}\leq 1-\alpha+(n+1)^{-1}.  
\end{align}
\gr{We refer to $\alpha$ as the {\it error} of the prediction set.} %

To illustrate our approach, as a thought experiment, assume that $X_{n+1}= X$, a data point at which an observation had already been taken, so that $X \in \{ X_1, \ldots , X_n\}$, and assume that its binary conformal prediction set is $\{0,1\}$, the widest it can be. Assume that the  previously observed instance of $X$ had $Y=1$ associated with it. Then $S(X, Y=1) \le S_{i:n}$ as 1 is in the prediction set for $X$. If now we could add $(X_{n+1}, Y=0)$ to the data set, then, knowing that $S(X, Y=0) \le S_{i:n}$ as 0 is also in the prediction set for $X$,
unless $i_{n+1} = i_n +1$, we have that $S_{i:n+1} = \max\{S(X, Y=0), S_{i-1:n} \}$, and thus the size of the conformal prediction \gr{set is no larger than the original set, but may be}  reduced, \gr{reflecting increased precision of the prediction.}.

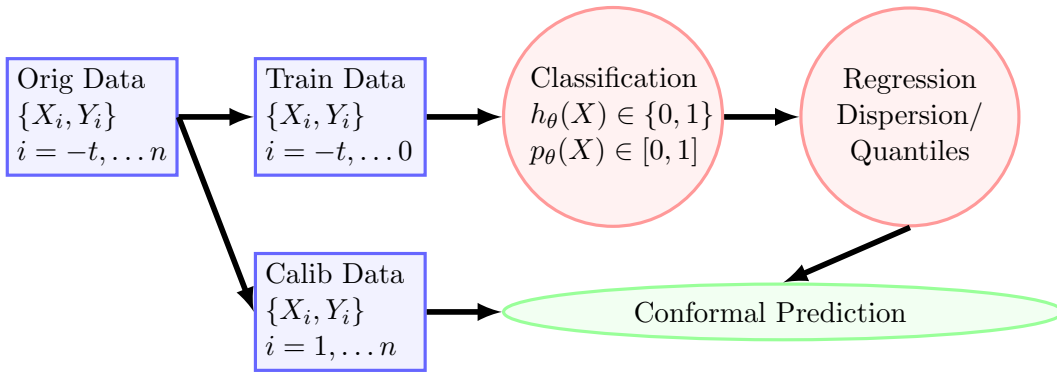
\begin{figure}
\begin{center}
\begin{tikzpicture}[
datanode/.style={rectangle, draw=blue!60, fill=blue!5, very thick, minimum size=10mm},
modelnode/.style={circle, draw=red!40, fill=red!5, very thick, minimum size=1mm},
conformnode/.style={ellipse, draw=green!40, fill=green!5, very thick, minimum size=1mm},
]
\node[datanode]        (origData)        {\begin{minipage}{2cm} Orig Data \\ $\{X_i,Y_i\}$ \\ $i=-t,\dots n$\end{minipage}};
\node[datanode]        (trainData)       [right=of origData] {\begin{minipage}{2cm} Train Data \\ $\{X_i,Y_i\}$ \\ $i=-t,\dots 0$\end{minipage}};
\node[datanode]        (calibData)       [right=of origData,below=of trainData] {\begin{minipage}{2cm} Calib Data \\ $\{X_i,Y_i\}$ \\ $i=1,\dots n$\end{minipage}};
\node[modelnode]        (classModel)       [right=of trainData] {\begin{minipage}{2.2cm} Classification
\\ \mbox{$h_\theta(X) \in \{0,1\}$}
\\ \mbox{$p_\theta(X) \in [0,1]$}\end{minipage}};
\node[modelnode]        (regModel)       [right=of classModel] {\begin{minipage}{2.2cm}
\begin{center}
Regression \\ Dispersion/ \\ Quantiles
\end{center}
\end{minipage}};
\node[conformnode]        (conform)       [below=of regModel, right=of calibData] {\begin{minipage}{5cm}
\ \ \ \  Conformal Prediction
\end{minipage}};

\draw[-latex, line width=0.751mm] (origData.east) -- (trainData.west);
\draw[-latex, line width=0.751mm] (origData.east) -- (calibData.west);
\draw[-latex, line width=0.751mm] (trainData.east) -- (classModel.west);
\draw[-latex, line width=0.751mm] (classModel.east) -- (regModel.west);
\draw[-latex, line width=0.751mm] (regModel.south) -- (conform.north);
\draw[-latex, line width=0.751mm] (calibData.east) -- (conform.west);
\end{tikzpicture}
\end{center}
\caption{An overview of how %
the conformal prediction intervals  $C_\alpha(X)$ are derived.} 
\label{fig:intervals}
\end{figure}

For binary $0-1$ classification these prediction sets provide a coarse 
quantification of uncertainty, as the only possible prediction sets are
$\{0\}, \{0,1\}, \{1\}$. To provide a fine grained determination of 
uncertainty we use the probabilities \gr{$p_\theta$} as determined by a classification model to create a regression model for the 
classification probability $p\in [0,1]$. %
We then  \gr{employ locally weighted conformal prediction, as detailed below,} %
using as scores the residuals of the regression model, weighted by their dispersion, to obtain a prediction interval for the class probability.
The pipeline for both, the binary classification problem and the probability prediction problem, is illustrated in Fig.~\ref{fig:intervals}

\paragraph{Locally Weighted Conformal Predictors}

An issue with standard conformal prediction is that the coverage guarantee \eqref{eq:covguarantee} is global; there may be subpopulations  which are considerably undercovered, see for example \cite{hore2025conformal}. One possibility to address this issue is to employ a conformity score function which includes a measure of the local variability. Here we use %
the Locally Weighted Conformal Predictor (LWCP) of \cite{lei_distribution-free_2018}, which provides a prediction interval for regression modelling  based on the locally weighted residuals 
$R_i= {| Y_i - \hat{\mu}(X_i)|}/{\hat{\rho}(X_i)}$,
where $(X_i, Y_i)$ is observed, $\hat{\mu}$ is a mean estimator, and $\hat{\rho}$ is the mean absolute deviation (MAD) measure of dispersion. We fit $\hat{\mu}$ and $\hat{\rho}$ sequentially. 
For an observed $X$ the locally weighted residual scores yield the conformal prediction interval  
\begin{equation} \label{eq:lwcp} \gr{C_\alpha (X) = }  [\hat{\mu}(X) - \hat{\rho}(X) d_\alpha, \hat{\mu}(X) + \hat{\rho}(X) d_\alpha]\end{equation} for $Y$, where $d_\alpha$ is derived from the calibration set and the 
desired coverage  $1- \alpha$. %
This prediction interval provides a non-negative estimate of uncertainty from its width, 
but it  symmetrical by construction, which may not be the best choice %
near the edges of the estimation range. \gr{An alternative is given for example by conformalized quantile regression.}

\paragraph{Conformalized Quantile Regression}

Conformalized quantile regression (CQR), proposed in \cite{romano_conformalized_2019} computes a prediction interval from a two regression models for the lower and upper quantile. Given the training samples $\mathcal{\gr{T}}$ a CQR constructs a prediction interval for a test point $X$ that %
contain\gr{s} the response variable $Y$ with probability $(1-\alpha)$.
To construct the prediction interval $C_\alpha(X)$ the training data is randomly split into parts $\mathcal{D}_\textrm{fit}$ and $\mathcal{D}_\textrm{calib}$. The first split $\mathcal{D}_\textrm{fit}$ is used to train two quantile regression models to predict the upper and lower quantiles, based on the pinball loss \citep{chung_beyond_nodate}. These models are then used to compute conformity scores for the samples in the second split $\mathcal{D}_\textrm{calib}$, and to obtain a prediction interval for the new sample. 
The upper and lower quantile estimates can 
accommodate the edges of the range required in the regression model. 
However, 
{typically} the upper and lower quantile models are computed independently, which can result in quantile crossing, and a {negative 
prediction interval
width.}

\medskip 
These conformal prediction intervals are a key ingredient in our counterfactual generation method. In contrast to the thought experiment, however, we do not leave it to chance whether or not we observe a suitable counterfactual instance. Instead, for a given point $X$, we choose a counterfactual according to a loss function which takes  the width of the conformal prediction interval as well as proximity between samples into account. %

\subsection{Distance between samples}
\label{sec:distance}

To %
measure the proximity between %
two samples $X'$ and $X$, with the first $m$ features continuous, and the remaining $p-m$ features categorical,  we use the weighted Gower distance %
\begin{equation}
    L_{\textrm{dist}}(X', X) = \frac{1}{m}\left(\sum_{j=1}^m \frac{|X'_j-X_j|}{R_j}\right)+\frac{1}{p-m}\sum_{j=m}^p \left(\frac{1-\delta_{X'_j,X_j}}{|S_j|}\right)
\end{equation}
where the feature components are denoted by the subscript $j$, $\delta_{X'_j,X_j}=1$ if the two categorical features $X'_j$ and $X_j$ are the same, and $0$ otherwise,  $R_j$ is the range of the continuous feature $j$, and $|S_j|$ is the number of categories for the categorical feature $j$. 
If there are no categorical features,  %
this distance reduces to the $L_1$ distance,  typically used in counterfactual generation, which induces sparsity \gr{with respect to the number of features which are modified to create the counterfactual}. %

\subsection{Overall individualised counterfactual framework}
\label{sec:framework}

To construct our individualised conformal prediction interval counterfactual (CPICF), we combine the ideas of the Sections~\ref{sec:modelKnowledge}, \ref{sec:conformal},
and \ref{sec:distance} into one overall framework. Suppose the example is $X$ and the required counterfactual is $X'$. The aim is to generate a counterfactual \gr{example} at \gr{a} point close to $X$ where the model predictions are most uncertain.

Recall that the \gr{entity} (a bank, say, from which a loan is sought) has the classifier $h_\theta$, whereas individual $k$ has classifier $\hat{h}_{\theta_k}$. A counterfactual must change the predicted class of the classifier of the organisation; thus we require that $h_\theta(X)\neq h_\theta(X')$.  
Further, we require that the counterfactual is close to $X$; thus we wish to minimise $L_{dist}(X,X')$. %
Using this loss alone we would obtain the following optimisation problem:  
\begin{equation}\label{eq:optimorig}
\underset{h_\theta(X)\neq h_\theta(X')}{\textrm{argmin}}L_\textrm{dist}(X, X'). 
\end{equation}

\noindent
This is effectively the standard formulation for counterfactuals in machine learning without adjustments for validity/realism etc.  In this paper, we wish to produce a counterfactual which also accounts for the knowledge of the individual $k$,  which we model \gr{by}  $\hat{h}_{\theta_{k}}$. As the individual in question only has access to {a limited set of points $\Tk$ with their corresponding probabilities \gr{(or scores)}},
we %
represent the uncertainty of the individual $k$  
using  
the width $C_\alpha(X)$ of the prediction  interval \eqref{eq:lwcp} 
through %
\[L^{(\Tk)}_{\textrm{info}}(X) = \frac{1}{C_\alpha(X)}.\] 
Adding this additional consideration to \eqref{eq:optimorig} gives
\begin{equation}
\underset{h_\theta(X)\neq h_\theta(X')}{\textrm{argmin}}\left( L^{(\Tk)}_\textrm{info}(X') + \lambda L_\textrm{dist}(X, X') \right)
\end{equation}
\noindent
where the %
\gr{scalar} $\lambda$ tunes the importance of the two components in the loss. For large values of $\lambda$, the second term dominates,  giving the original form of a counterfactual. For $\lambda=0$, the counterfactuals will find the point of largest uncertainty which will change the classifier. %
For a given $X$ the counterfactual $X'=X'(X)$ is completely determined by this optimisation (\gr{with a deterministic tie breaking rule}). \gr{The counterfactual construction is carried out by the entity and tailored to the information that individual $k$ has available, as follows. The entity constructs a  LWCP interval  by using a regression model for the probabilities $p_\theta(X_i)$, with $X_i$ in the calibration data set, against the estimated probabilities $p_{\theta_k} (X_i)$, available to individual $k$. Similarly, the entity uses a regression model to assess the dispersion of the model around a particular data point, yielding $\hat{\rho}$. 
}
The full algorithm for %
is summarised in Algorithm 1. 

\begin{algorithm}
\rule{\textwidth}{1.5pt}
\caption{Conformal Prediction Interval Counterfactual (CPICF)}\label{alg:CFCP}
\vspace{-1ex}
\rule{\textwidth}{1.5pt}
\vspace{2ex}
\hspace{1em}
\begin{minipage}{0.95\textwidth}    
\DontPrintSemicolon
\textbf{\hspace{-0.5em}Input:\\}
Data set $\D=(X,Y)_{i=-t}^n$\\
Query instance for counterfactual $X$\\
Binary classifier $\hat{Y} = h_\theta(X)$\\
Class probability $p_\theta(X) = P(Y=1| X, \theta)$\;

\textbf{\hspace{-0.5em}Process: }\\
Compute regression targets $Y_i = p_\theta(X_i)$ for $X_i \in \Tk$ using class probabilities from $h_\theta$\\
Train a regression model $%
{p}_{\theta_k} (x) $ using  $\Tk$ \\
Train a regression model $\hat{\rho}$ using $X_i\in \Tk$ with the target of $|Y_i-p_{\theta_k}(X_i)|$\\
Compute the LWCP sets using \eqref{eq:lwcp}
\\
Compute counterfactual $X'$ to from query instance $X$ by constrained minimisation of
$$\underset{h_\theta(X)\neq h_\theta(X')}{\textrm{argmin}}\left( L^{(\mathcal{D}_k)}_\textrm{info}(X') + \lambda L_\textrm{dist}(X, X') \right)$$
\textbf{\hspace{-0.7em}Output:}
Counterfactual $X'$  for query instance point $X$.
\end{minipage}
\hrule
\end{algorithm}
\subsection{Implementation}

The framework for CPICF generation detailed in Section~\ref{sec:framework} was implemented for a gradient boosted  model (XGboost; XGB for short, from \cite{XGBcitation}) to enable its evaluation. This implementation was used for training the classification model $h_\theta$, and the individual's regression model $p_{\theta_k}$. XGB models were used for the input into the conformal prediction intervals, for training the dispersion models for LWCP, and for the quantile regression models in CQR. 
The PUNCC python library was used to %
compute the prediction intervals from the regression models and a calibration set \citep{mendil_puncc_2023} (details of the division between training and calibration sets are given in Sec. \ref{sec:exp}).

To generate a CPICF the constrained minimisation problem was solved using the multi-objective optimisation in python (pymoo) genetic optimisation algorithm \citep{blank_pymoo_2020}. This method is particularly useful for mixed data types, as is often found in tabular data. %
The \texttt{pymoo} genetic algorithm parameters were set to $50$ evaluations and a population 
size of $20$.

\subsection{Locally evaluating the  counterfactual framework}
\label{sec:deltaevaluationframework}
There are multiple ways to 
evaluate a counterfactual ranging from quantitative statistics to qualitative measures (e.g., user studies). Each highlights the different properties that we may wish a counterfactual to have see for example \cite{guidotti_counterfactual_2024} for a discussion of some possible measures.
In this work, we are interested in evaluating the individual effect of a counterfactual on the individual's knowledge of the underlying model. Thus we evaluate the counterfactual by examining how the model probabilities around the requested instance $X$ change with the addition of the counterfactual.  
This can be seen as a generalisation of one of the evaluations in \cite{mothilal_explaining_2020}, which measures the ability of a $k$-nearest neighbour classifier trained on the counterfactual examples (and the original input) to classify instances at different distances from the original point.
In this generalisation we directly consider the knowledge of the individual, use a more complex classifier and explore the probabilities rather than the classes as described below.  

The proximity of the model prediction probabilities to the underlying model quantifies the effect of the counterfactual. \gr{We perturb $X$ by $\omega$ and record the}  change (absolute deviation) at $\omega$ from the query instance $X$ \gr{with observed $Y$},
\begin{equation}
\Delta_k(X;\omega) = | p_\theta(X+\omega) - p_{\theta_k}(X+\omega)|.    
\end{equation}
After the addition of a counterfactual \gr{$(X', Y')=(X'(X), Y')$, with $Y'$ the opposite of $Y$}, the absolute deviation in prediction probabilities \gr{with the perturbation $\omega$} is
\begin{equation}
\Delta_{k;X'}(X+\omega) = | p_\theta(X+\omega) - p_{\theta_{k;X'}}(X+\omega)|.   
\end{equation}
\gr{Here $p_{\theta_{k;X'}}$denotes the classifier for individual $k$ trained on ${\mathcal{D}}^{(k)} \cup \{ (X',Y')\}.$}
By examining this change \gr{in a region $\Omega={\Omega(X)}$ in feature space centred around the query instance $X$}  %
we can quantify the improvement of the model:
\begin{equation}
\Delta (X) = 
\frac{1}{| \Omega^{(g)} (X) |}\sum_{\omega\in \Omega^{(g)}(X) }\left[\Delta_{k;X'} (X+\omega) -\Delta_{k} (X) \right] 
\label{eq:deltaX}
\end{equation}
where $\Omega^{(g)} =\Omega^{(g)} (X)$ \gr{is a regular grid of points}  in $\Omega$ \gr{which we use to approximate} $\Omega=\gr{\Omega(X)}$.
Then $\Delta < 0$ demonstrates that the counterfactual has moved the prediction probabilities closer to the ``oracle'' model \gr{$p_\theta$} around the query instance $X$.
We summarise the performance of CPICFs by considering the distribution of $\Delta$ over the a set of sample points. 
\noindent

\section{Experiments and discussion} \label{sec:exp}

Two datasets were investigated to assess the counterfactual generation using this
technique 
-- a synthetic classification problem from sci-kit learn \cite{scikit-learn} based on Gaussian clusters around the vertices of a hypercube (the \textit{hypercube dataset}), see \cite{guyon_design_2003} for full details, and a synthetic fraud detection data set with both continuous and categorical data types (Tabformer)
\citep{padhi_tabular_2021}. The hypercube dataset is low dimensional, and is easier to visualise, whilst the fraud  data set is representative of real world tabular data, with mixed data types.

\subsection{The hypercube dataset}

\begin{figure}
\includegraphics[width = \textwidth]{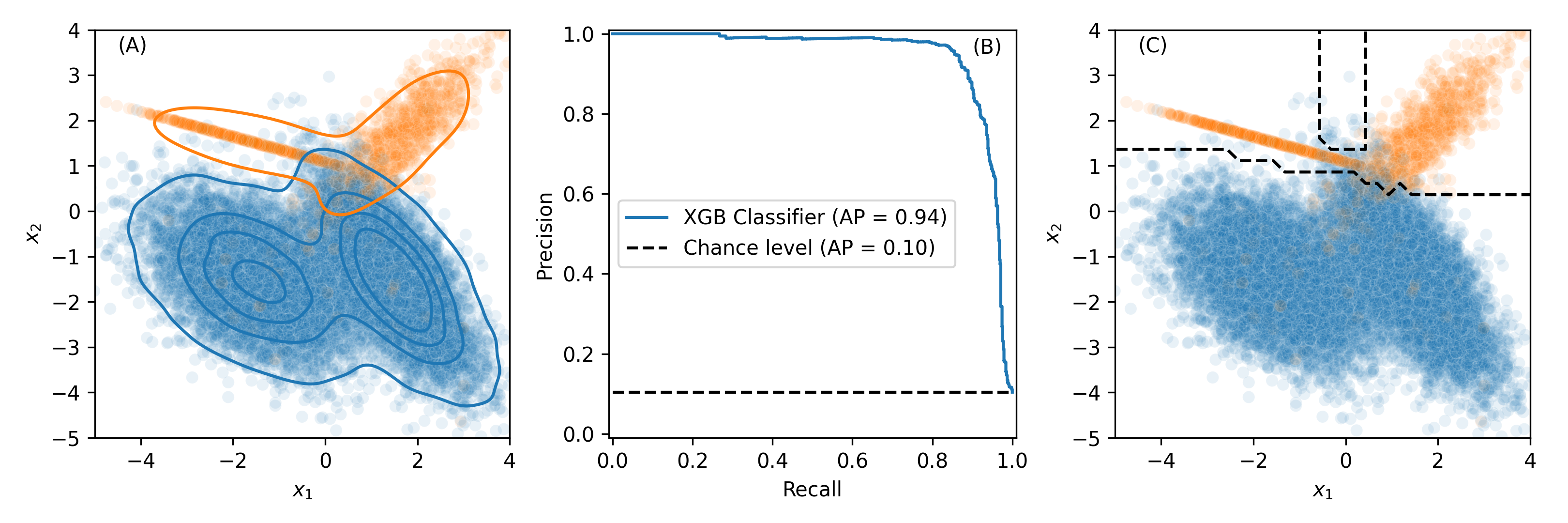}
    \caption{%
    (A) The original classification problem with the points coloured according to their classification (orange and blue), %
    (B) XGB classifier precision-recall curve and average precision (AP) when trained on $60\%$ of the data and tested on a $20\%$ held out fraction, 
    and (C) the decision boundary of the classifier (dashed line). } 
    \label{fig:classificationproblem}
\end{figure}

A synthetic binary classification hypercube dataset was created, with $30000$ samples, and a significant class imbalance of $90\%$ to $10\%$ to make the problem %
challenging. Fig.~\ref{fig:classificationproblem} illustrates the feature space for the two classes, together with the performance of the XGB classifier ($h_\theta$) and the decision boundary. To train the classifier the dataset was split into a training and cross validation set ($60\%$), calibration set ($20\%$) and a test set ($20\%$). Cross validation was used for hyperparameter selection,  see 
Appendix \ref{sec:appendixmodelparameters}.

\begin{figure}
\includegraphics[width = \textwidth]{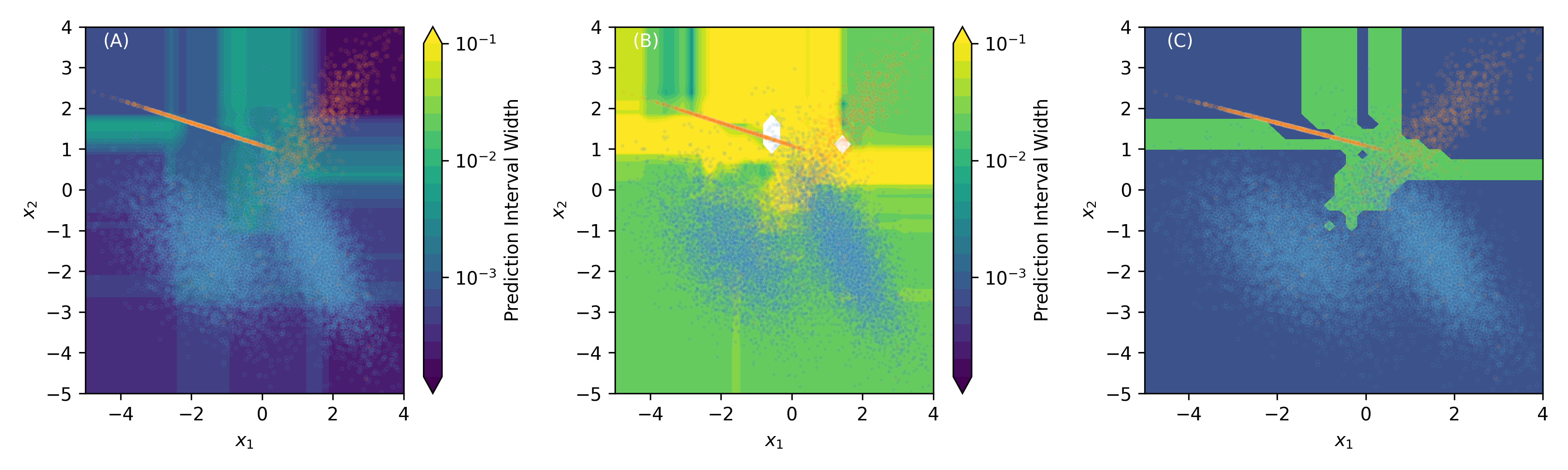}
\caption{%
\gr{T}he prediction interval width (darker colour corresponds to smaller prediction interval) for conformal predictors with $\alpha = \gr{0.2}$,
based on the same training set,  %
using (A) LWCP or (B) CQR, and (C) size of %
the conformal prediction set based on direct binary classification %
\gr{T}he full training data $\mathcal{T}$ is overlaid (NB: white regions in (B) correspond to quantile crossing with %
\gr{conflicting (negative)} prediction intervals). 
}
\label{fig:2Dconformalintervals}
\end{figure}

Fig.~\ref{fig:2Dconformalintervals} illustrates the width of the conformal prediction interval for the hypercube dataset for both LWCP and CQR. The prediction intervals are computed for an error %
of
$\alpha = 0.2$ 
for {each} conformal predictor. 
Additionally the more coarse grained measure of prediction set size computed from a conformal classifier is shown. These 
examples confirm the intuition that prediction uncertainty is larger along the decision boundary. Whilst the 
techniques are broadly similar, the \gr{LWCP and CQR} prediction interval techniques provide a more fine-grained measurement of the local uncertainty. Note that the CQR intervals occasionally cross (denoted white in 
Fig.~\ref{fig:2Dconformalintervals}) 
- see Sec.~\ref{sec:conformal} for details.

\begin{figure}
\includegraphics[width = \textwidth]{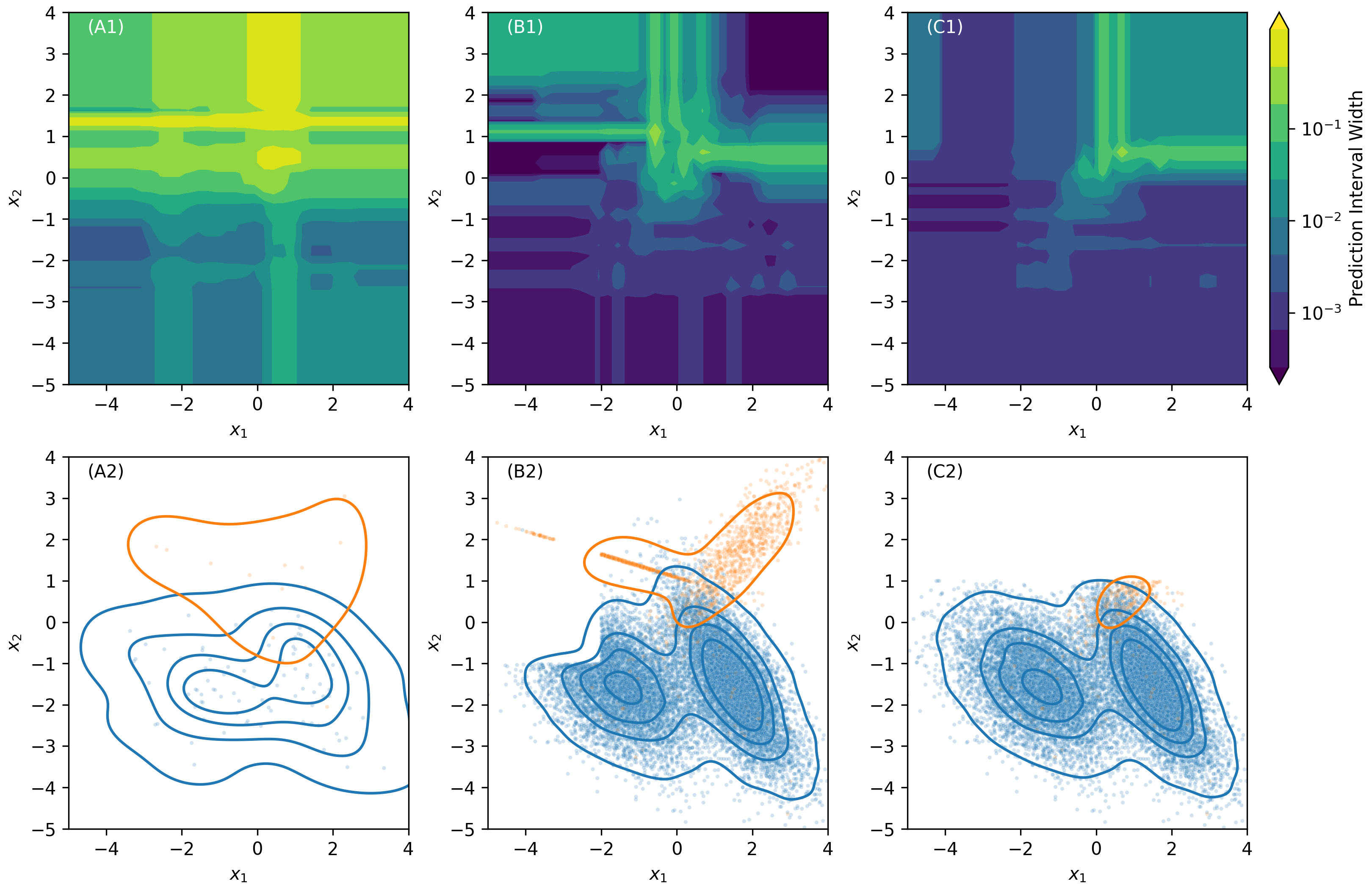}
    \caption{%
    Prediction intervals using LWCP  for different training datasets, and the same calibration data set. (A1):  for the sparse training data in (A2), (B1-2): ablating an area on the left, and (C1-2):  ablating all points for $x_2>1$.}
    \label{fig:2Dvarywidth}
\end{figure}

The effect of modifying the training set on the prediction interval widths is shown in Fig.~\ref{fig:2Dvarywidth}. Sparse datasets tend to have larger prediction intervals, particularly in regions of feature space where the minority class is present (A1). The conformal prediction interval is wider in regions where there are very few data points, as might be expected intuitively.

\begin{figure}
\includegraphics[width = \textwidth]{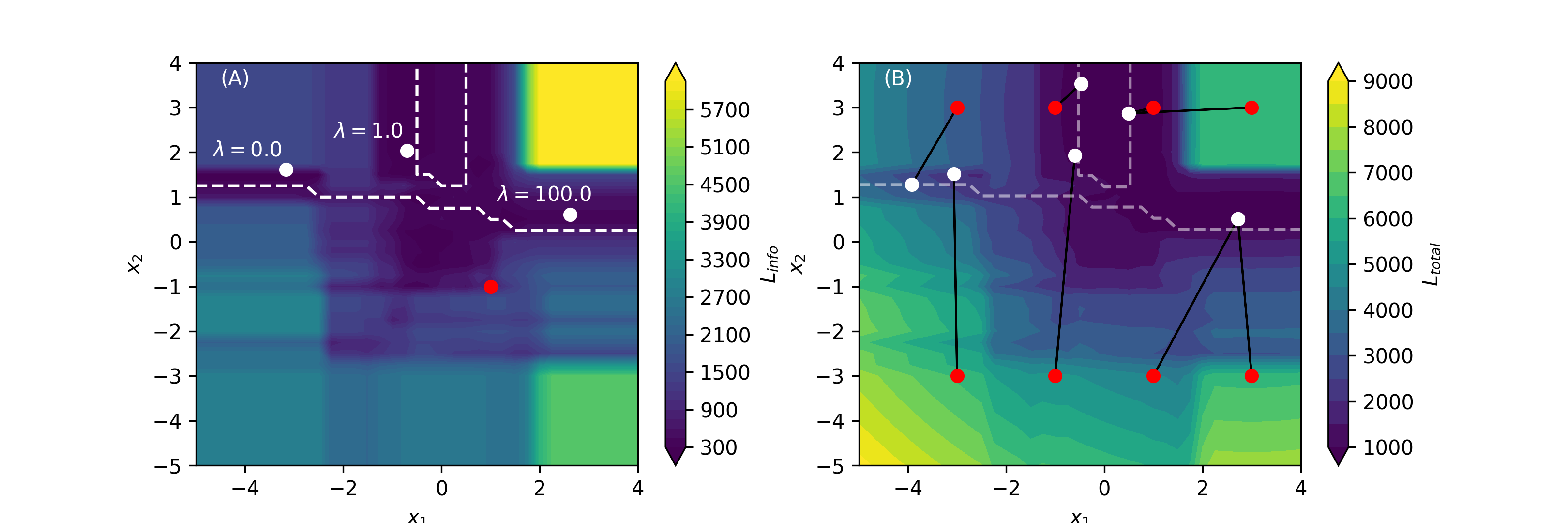}
    \caption{%
    (A) The impact of the parameter $\lambda$ on the counterfactual selection, showing counterfactual instances for different $\lambda$ values on the $L_\textrm{info}$ objective function and (B) CPICFs for different starting instances for the same $L_\textrm{total}$ objective function.}
    \label{fig:2DCFinstance}
\end{figure}

Some CPICFs for the hypercube dataset are shown in Fig.~\ref{fig:2DCFinstance}. Increasing the value of $\lambda$ moves the CPICFs (white in the figure) closer to the original instance (red). This figure highlights the potential lack of diverse counterfactuals derived from this technique -- if the prediction intervals have one very deep area of uncertainty, then they will tend to derive counterfactuals from this region. This might be informative for a few instances of counterfactuals, but subsequent examples will severely lack sample diversity. 

\subsubsection{Change in local prediction probability}\label{sec:individualCFA}

The evaluation framework from 
Sec.  \ref{sec:deltaevaluationframework}
which measures the improvement in an individuals knowledge with a single counterfactual
was applied to the hypercube dataset. We randomly selected a data subset containing $100$ sample points to represent the individual's knowledge $\Tk$, and then generated $100$ independent 
CPICFs, \gr{each} based on a randomly selected training point as the query instance, using sampling with replacement, with different genetic algorithm seeds, but with the same starting classifier $h_{\theta_k}$ to compute the conformal prediction intervals. This is  repeated for $7$ independent realisations (random seeds) of the data set. Counterfactuals were generated with different objective function parameter $\lambda$, as well as with a constant objective function, chosen to select a random counterfactual. \gr{The entry ``Gower distance'' refers to $\lambda = 10^5$.}

\begin{figure}
        \includegraphics[width = \textwidth]%
        {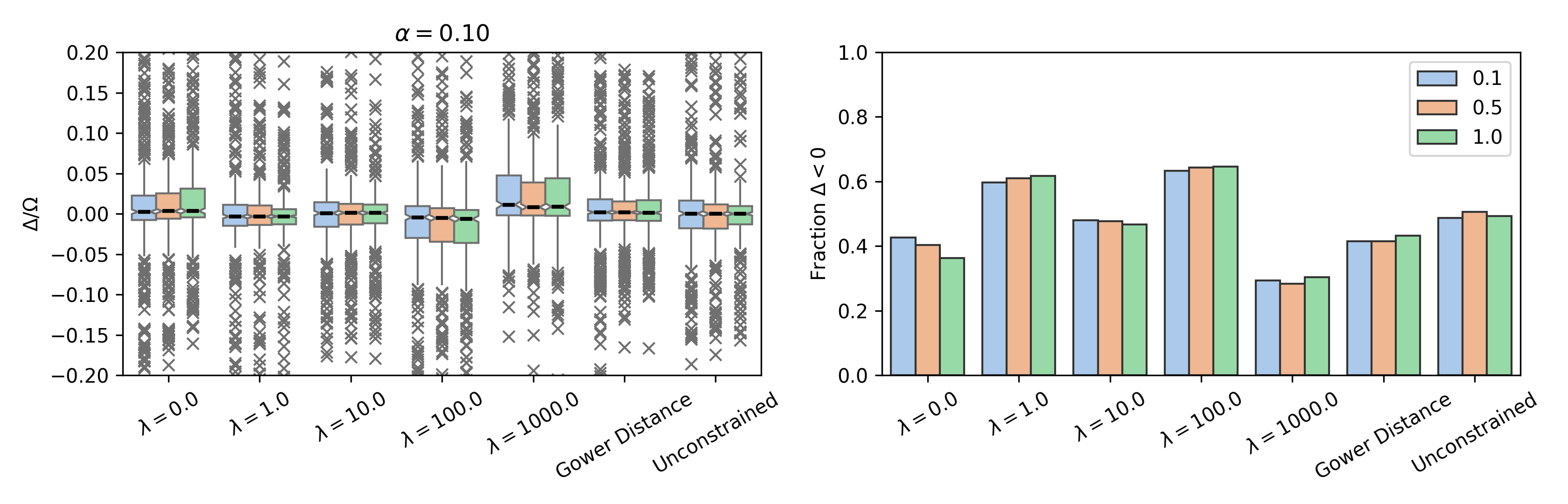}
    \caption{A boxplot (left) of the change in the MAD prediction probabilities between with 
    and without the counterfactual in the training set. 
    $100$ observations are sampled without replacement in equal proportion from each class in the training data. Each point corresponds to the $\Delta\gr{(X)}$ following an $X'$ obtained from $X$ in the training data.
    The $\Omega\gr{(X)}$ used for averaging is a square around $X$ {of side length $0.1$, $0.5$ and $1.0$ represented by blue, orange and green bars respectively,} 
     and the error %
    is $\alpha = 0.1$. The fraction of the $\Delta < 0 $ is shown on the right, illustrating that for \gr{$\lambda =1$ and} $\lambda \gr{=100}$ there is an improvement over using only the length of the conformal prediction interval ($\lambda =0$), or using (mainly)  Gower distance \gr{($\lambda = 10^5$)}, \gr{or using general counterfactuals (``unconstrained'')}. %
    } 
    \label{fig:localCFimpact}
\end{figure}

Fig.~\ref{fig:localCFimpact} %
\gr{shows $\Delta(X)$ from \eqref{eq:deltaX}} for $\Omega\gr{(X)}$ \gr{a square of} side $0.1, 0.5$ and $1.0$. For $\lambda =1$ or $100$, prediction probabilities for the classifier with single additional counterfactual in the training data typically move closer to those oracle classifier $p_\theta$; \gr{the fraction of negative $\Delta$ is substantially larger than $50 \%$.} %
\gr{Moreover, creating counterfactuals only based on the width of the conformal prediction set ($\lambda =0$), or \gr{mainly} based on the Gower distance (a large value of $\lambda = 10^5$), or only based on being a counterfactual (``unconstrained'') on average does not improve the prediction. We also note that the size of $\Omega(X)$ does not seem to have much of an effect.}
\gr{Results for} $\alpha = 0.2$ are included in Appendix \ref{sec:Supplementary}, \gr{with a similar finding, except that the choice of $\lambda$ would then be 10. Investigating possible relationships between $\alpha$ and the best $\lambda$ will be part of future work}.
\subsubsection{The effect of conformal prediction interval counterfactuals
}

To %
\gr{further} 
assess whether CPICFs are informative, {rather than considering the impact of a single counterfactual,} 
\gr{e}ach training data point was augmented with $0, 1$ or $4$ of counterfactuals in a batch\gr{, using $\alpha = 0.1$}, and then the performance of an XGB classification model was evaluated on the test set. The results of these calculations on the hypercube dataset are shown in Table~\ref{tab:2Daugmentation}. 

\begin{table}[ht]
\begin{center}
\begin{tabular}{ccccc}
\hline
Sample Points & Aug. per sample & Average Precision & F1 Score & ROC AUC \\
\hline \hline 
50 & 0 & $0.79\pm0.05$&  $0.76\pm0.14$ & $0.96\pm 0.01$  \\
\hline
50 & 1 &$0.88\pm 0.01$& $0.85\pm 0.02$ & $0.96\pm 0.01$\\
\hline
50 & 4 &$0.86\pm 0.04$&  $0.83 \pm 0.01$ & $0.96\pm 0.01$ \\
\hline
\hline
500 & 0 &$0.87\pm 0.01$ & $0.87\pm 0.01$ & $0.96 \pm 0.002$\\
\hline
500 & 1 &$0.92\pm 0.01$& $0.90\pm 0.01$ & $0.97 \pm 0.001$\\
\hline
500 & 4&$0.91\pm 0.01$&$0.90\pm 0.01$ & $0.97 \pm 0.001$\\
\hline
\end{tabular}
\end{center}
\caption{The performance of a classifier trained with increasing levels of data augmentation from counterfactual examples derived with $\lambda = 1000$. The classifier 
trained on
$18000$ data points achieved an average precision of $0.94$, an F1 score of $0.90$, and a ROC AUC of $0.98$. The error estimates are based on 
$3$ replicates.} %
    \label{tab:2Daugmentation}
\end{table}
For a small number of data points the counterfactual augmentation improve both the average precision and the F1 score, demonstrating that the counterfactual convey information to their recipient. The effectiveness of the augmentations diminished as the number of augmentations increases, potentially indicating a lack of diversity in the batch of counterfactuals. 
A classifier with a larger number (500) of training points improves its average precision and F1 score to a lesser degree; this model was already close to the performance ceiling of a model trained on a much larger data set. 

%
\subsection{A Transaction Fraud Dataset}

Tabformer \citep{padhi_tabular_2021} is a synthetic multivariate tabular time series transaction fraud dataset.  It has $24$ million transactions from $20,000$ users, with $12$ features and a binary label \texttt{isFraud?} classifying each transaction. It contains continuous features such as \texttt{Amount}, categorical features such as \texttt{Use Chip}, and other categorical features with a high cardinality such as \texttt{Merchant City}. Tabformer has many of the characteristics that make detecting transaction fraud challenging,  including a class imbalance with $<1\%$ fraudulent transactions. The dataset exhibits distributional shift,  for example as new card types become available, or transaction habits change from chip and pin to online transactions.

\gr{To assess the effect of conformal prediction interval counterfactuals, for this data set }, 
a temporal split %
was used. A three year period from 02-04-2015 to 02-04-2018 was used for the training set. This was further divided into $75\%$ for training and cross validation, and $25\%$ for the conformal predictor calibration set. A $1$ year period after 02-04-2018 was used for the test data set. 

An XGB classifier ($h_{\theta_{(\D)}}$) was then trained, and used to provide the targets for %
the LWCP. As the data set is still imbalanced (with approximately $7500$ fraud transactions out of over $5$ million) the majority class was randomly undersampled to balance the classes. The XGB model was trained with continuous representations for the Time, and Amount features, one hot encoding for low cardinality categorical features, and target encoding for the high cardinality features such as Zip. After hyperparameter tuning the resulting model had an average precision of $0.65$, ROC AUC of $0.994$ and an F1 Score of $0.11$ on the $1$ year of hold out data in the test data set.%

\gr{The dispersion model for $\hat{\rho}$ from Algorithm \ref{alg:CFCP}, needed  for the LWCP method,}
was trained on a smaller dataset of $4000$ samples \gr{from the training set}, \gr{representing the {uncertainty in the} knowledge of an individual.} %
The derived model was used for computing CPICFs, with the parameters used given in Appendix \ref{sec:appendixmodelparameters}. A typical example of a counterfactual is given in Table~\ref{tab:tabformerCFexample}. This example shows that the modifications made to a particular query, whilst they may be informative, are not sparse in the \gr{number of} features that are changed.

\begin{table}[!ht]
\small
    \begin{tabular}{|p{2cm}|p{1cm}|p{1.25cm}|p{1cm}|p{1cm}|p{3.5cm}|p{2.5cm}|}
    \hline
       Orig/CPICF & Time &Amount&User&Zip&Merchant Name&Merchant City
       \\\hline
Orig &       13.17  &	39.63&1816&8270&-7051934310671642798&Woodbine
        \\\hline
CPICF &	  10.64   &  1635.89  &  1768  &  99163   & -264402457929754322 & Piney River
    \\\hline
    \end{tabular}
    
    \vspace{2mm}
    
    \begin{tabular}{|p{2cm}|p{2.5cm}|p{1.25cm}|p{2cm}|p{1cm}|p{2.0cm}|p{1.5cm}|}
   \hline 
       Orig/CPICF & Merchant State &
       Card&Use Chip&MCC&Errors?& Weekday    
       \\\hline
Orig & NJ        &0&Chip 
&5310&nan&3
    \\ \hline   
CPICF & FL&3&
Online 
&3256&Bad Zipcode&2
       \\\hline
    \end{tabular}
    \caption{An example of a non-fraudulent transaction (Orig), and a fraudulent CPICF derived from it for the Tabformer data set, with %
   \gr{error} $\alpha = 0.2$ and $\lambda = 10^4$.}
\label{tab:tabformerCFexample}
\end{table}

\begin{figure}[!htb]
\centering
\includegraphics[width = 0.8\textwidth]{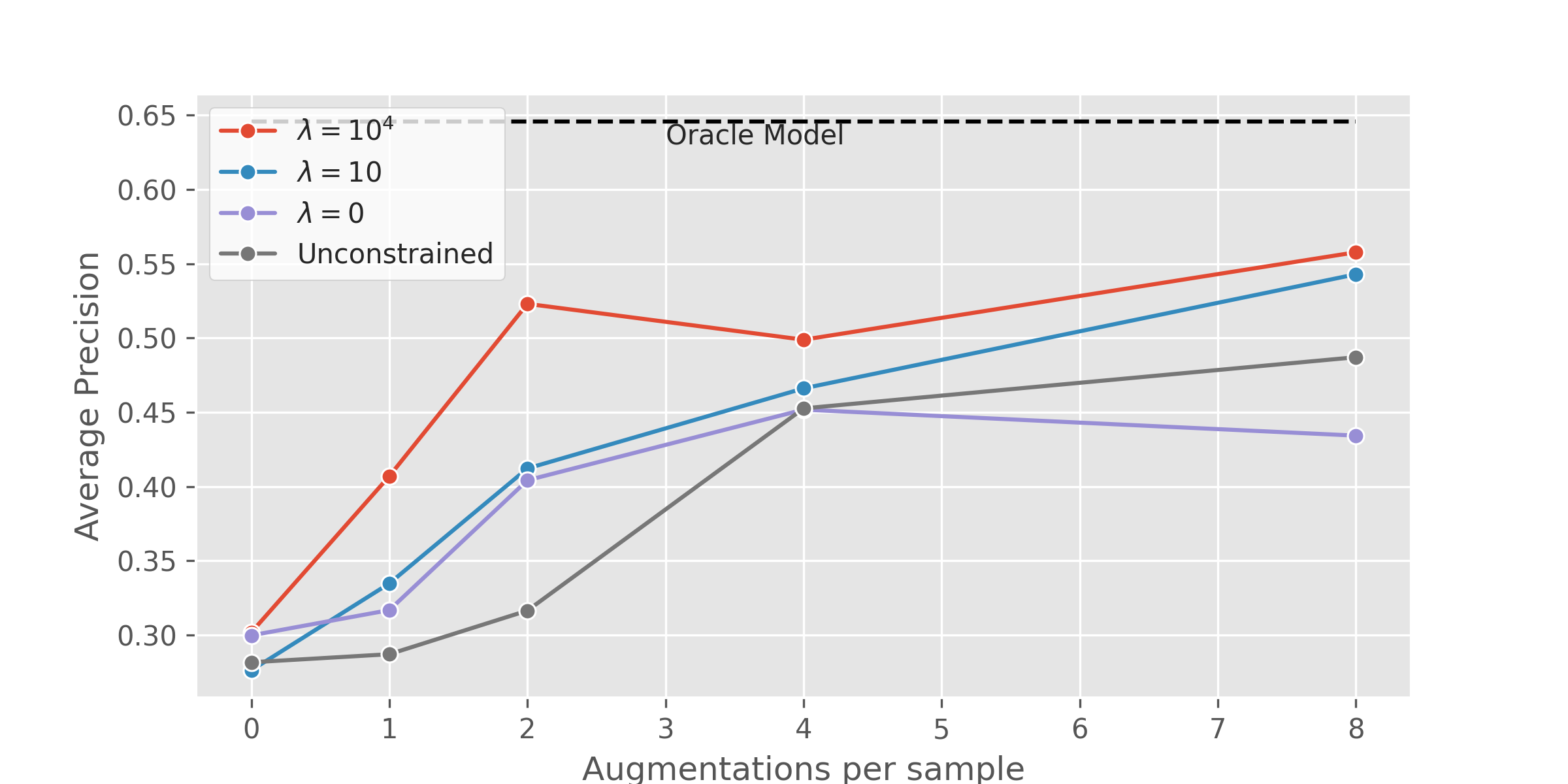}
    \caption{
    \gr{Average precision (area under the precision-recall curve) after} training augmented with \gr{CPICF} %
    examples %
    for the %
    Tabformer %
    dataset, %
    for $\lambda \in  \{0, 10,10^4\}$, \gr{$\alpha = 0.2$}, and an unconstrained counterfactual 
    with an alternate label but 
    selected without the distance or conformal loss terms. 
    }
    \label{fig:tabformeraugmentation}
\end{figure}

The effectiveness of the CPICFs was measured by examining their effect in data augmentation (Fig.~\ref{fig:tabformeraugmentation}). %
\gr{Similarly as for Figure \ref{fig:localCFimpact} 
we compare the performance of the CPICFs for $\lambda =0, 10, 1000,$ as well as unconstrained counterfactuals (random points with the alternate class). Here we take $\alpha = 0.2$.}

Adding the counterfactuals to the training set \gr{clearly} improves the model performance \gr{on average}, \gr{providing some evidence for the claim that} %
CPICFs \gr{can} provide a more informative augmentation when compared with unconstrained counterfactual augmentation. %

\section{Conclusion} \label{sec:concl}

\gr{Here we present a framework for a large entity (such as a bank) to derive counterfactual examples for an individual. The assumption is that the entity holds a large dataset and  a potentially sensitive classifier derived from the data, while the individual has only limited knowledge of the data set.}
\gr{In contrast to standard approaches in counterfactual generation which use a parametric estimator of the underlying distribution, and in contrast to standard classification methods which learn a classification boundary,}
to compute these counterfactuals, a \gr{(nonparametric)} conformal prediction set  is used to express the uncertainty of the \gr{individual}, who  has a smaller dataset, and a less clear understanding of the decision boundary. %
An objective function with two terms,  expressing the proximity of the counterfactual example  and the uncertainty of the individual \gr{with a relative weight factor $\lambda$}, is minimised, subject to the constraint of the counterfactual having the opposite classification. %
\gr{We call the {point that achieves this} minimum  a} conformal prediction interval counterfactual (CPICF). 

\gr{This counterfactual generation method is} %
evaluated by assessing the change in the predictive probabilities of the classification model \gr{of the individual after adding the counterfactual to the data set}, and \gr{by assessing its} potential for data augmentation \gr{through considering the average precision, F1 score, and ROC AUC. In our synthetic hypercube example as well as in the Tabformer data set, CPICFs can improve the prediction given an appropriate choice of the  parameter $\lambda$ {which trades off the importance of the distance the counterfactual is from the original point and the level of classifier uncertainty}}. 

\gr{While our experiments validate the CPICF approach, for a practical deployment there are more experiments which could be interesting to carry out, such as whether there is a variation of the CPICF performance in different regions of the  feature space. More parameter settings could be studied, and comparisons with more counterfactual generation methods could be carried out. Moreover, assessing faithfulness or fidelity, or other utility measures often employed for assessing counterfactuals, could be of interest.}

\gr{Next we point out several possible improvements to the algorithm.} Typically, conformal prediction techniques are model agnostic, but are data inefficient, as they require an additional data partition for the calibration set. \cite{stutz_learning_2022} have developed a more efficient end-to-end training procedure that incorporates the calibration set during model training. Incorporation of this procedure could be %
\gr{linked to} our pipeline as future work.  
\gr{Moreover, we} limited ourselves to locally weighted conformal prediction techniques to compute a positive size of the interval width, as conformal quantile regression can result in quantile crossing when each quantile is fitted independently. Concurrent quantile regression models    \cite{cannon_non-crossing_2018} %
might be used to obtain non-crossing quantile models which could be explored as part of future work.

\gr{Another line of future research relates to the theoretical underpinning of the CPICF generation.} \gr{Standard conformal prediction requires the input data to be exchangeable. Thus, once the counterfactual is added, the conformal prediction guarantees no longer hold. 
However, the CPICFs  $X' = X'(X)$ are generated \gr{in a particular way} based on prediction intervals which use the calibration data. %
While the CPICFs and the calibration data are not exchangeable, i}t may be possible to apply ideas from \cite{hore2025conformal} to randomise the intervals in order to obtain theoretical guarantees; investigating this will be part of future work.

\gr{Finally, as a different application, t}he quantification of model uncertainty by conformal prediction interval width might be useful in %
contexts where the challenge is to optimise the collection of data. For example, classification probabilities to quantify the information content of an observation is often used in optimising the collection of sensor information \citep{krajnik_life-long_2015}, such as building in occupancy models from sequential collections. However this prioritisation can lead to focusing on observations that are intrinsically difficult to predict (such as occupancy during a time of transition which will always be close to $0.5$, or more generally observations at a boundary), rather than where model uncertainty could be reduced.
Quantifying the uncertainty through a conformal prediction interval might lead to better sensor deployment.

\acks{The authors acknowledge the support of the UKRI Prosperity Partnership Scheme (FAIR) under EPSRC Grant EP/V056883/1 and the Alan Turing Institute, and the Bill \& Melinda Gates Foundation. GR is also funded in part by the UKRI EPSRC grants EP/T018445/1, EP/X002195/1 and  EP/Y028872/1.}
%

%
%

%
%

%
%
%
%
%
%
%

\newpage 
\appendix

\section{Model parameters}
\label{sec:appendixmodelparameters}

XGB hyperparameter tuning was conducted using cross validation 
with the following values:\\

\begin{tabular}{ll|ll}
Name & Values & Name & Values \\ \hline 
            n\_estimators & $[80,100,120,150,300]$ & 
            max\_depth & $[8,9,10,11,12]$ \\ 
            learning\_rate & $[0.01,0.05,0.1,0.20]$ & 
            min\_child\_weight & $[1,2,3,4,5]$ \\ 
            subsample & $[0.6,0.8,1.0]$ & 
            colsample\_bytree & $[0.6,0.8,1.0]$ \\
\end{tabular}

\section{Supplementary Counterfactual Assessment Figures}
\label{sec:Supplementary}

\begin{figure}[!h]
        \includegraphics[width = \textwidth]%
        {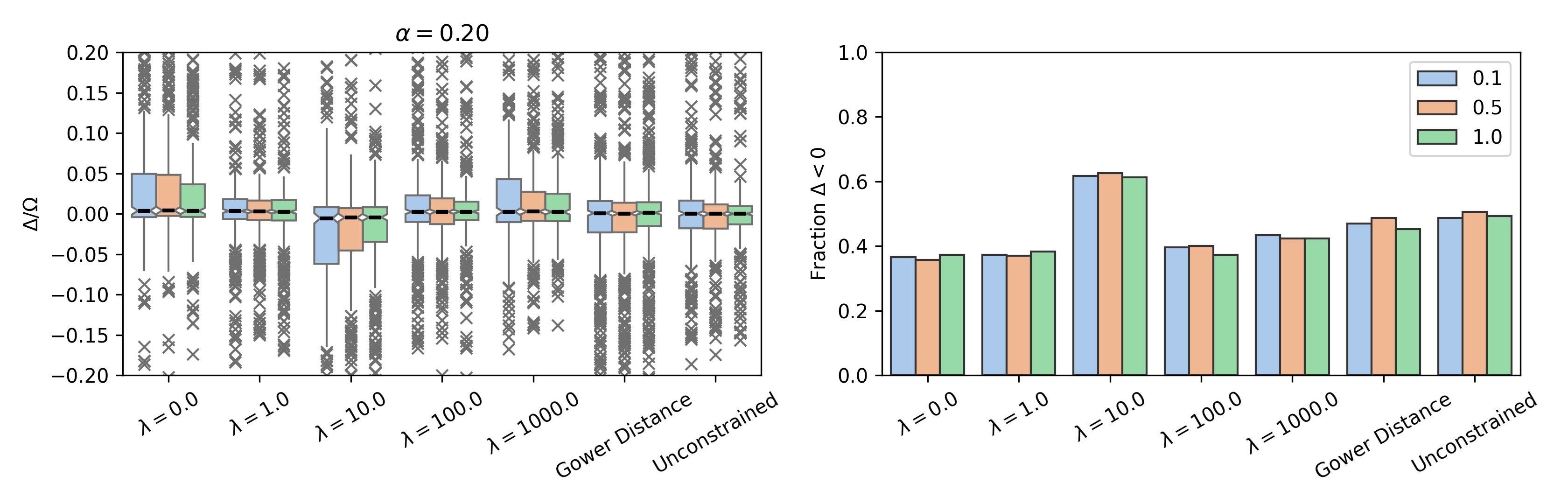}

    \caption{%
   The change in the local model prediction following the inclusion of a counterfactual in the training set for $100$ repetitions and $7$ independent realisations of the hypercube data set, with an error %
    of $\alpha = 0.2$
} 
\label{fig:alpha02}
\end{figure}
\gr{While Figure \ref{fig:localCFimpact} showed results for $\alpha = 0.1$,} %
Figure \ref{fig:alpha02} shows \gr{results for $\alpha = 0.2$ and illustrates} that there is a regime for $\lambda $ in which 
  there is an improvement over using mainly Gower distance ($\lambda =10^5$) or only the length of the conformal prediction interval ($\lambda =0$). However, now the improvement is seen for a  smaller value of $\lambda$ \gr{than for $\alpha =0.1$,} highlighting the need for careful calibration of $\lambda$.

\end{document}